\documentclass{article}
\usepackage{spconf,epsfig}

\usepackage{color}
\usepackage{amsfonts}
\usepackage{amssymb}
\usepackage{amsbsy} % Per a \boldsymbol
\usepackage{amsmath}
\usepackage{graphicx}
\usepackage{epstopdf}
\usepackage{amsthm}
\usepackage{bbm}

\newcommand{\param}{\mathbbm{z}}
\newcommand{\Param}{\mathbbm{Z}}

\newcommand{\tit}{Causal Inference in Geosciences with Kernel Sensitivity Maps}

\usepackage{array}
\newcolumntype{L}[1]{>{\raggedright\let\newline\\\arraybackslash\hspace{0pt}}m{#1}}
\newcolumntype{C}[1]{>{\centering\let\newline\\\arraybackslash\hspace{0pt}}m{#1}}
\newcolumntype{R}[1]{>{\raggedleft\let\newline\\\arraybackslash\hspace{0pt}}m{#1}}

\usepackage{pgffor} % for 
\usepackage{float} % avoid floats error latex
\usepackage{wrapfig}

\usepackage{amsmath}
\usepackage{amssymb}
\usepackage{amsthm}
\usepackage{algorithm}
\usepackage{algorithmic}
\usepackage{placeins}
\usepackage{multirow}
\usepackage{color}
\usepackage{url}
\usepackage{cite}
\usepackage{wrapfig,lipsum,booktabs}
\usepackage[font=small,labelfont=bf]{caption}
\usepackage{graphicx}
\usepackage{xcolor}
\usepackage[all]{xy}
\usepackage{hyperref}
\usepackage{adjustbox}
\hypersetup{
    bookmarks=true,         % show bookmarks bar?
    unicode=false,          % non-Latin characters in Acrobat’s bookmarks
    pdftoolbar=true,        % show Acrobat’s toolbar?
    pdfmenubar=true,        % show Acrobat’s menu?
    pdffitwindow=false,     % window fit to page when opened
    pdfstartview={FitH},    % fits the width of the page to the window
    pdftitle={\tit},    % title
    pdfauthor={Adrian Perez-Suay and Camps-Valls},     % author
    pdfsubject={\tit},   % subject of the document
    pdfcreator={Adrian Perez-Suay and Camps-Valls},   % creator of the document
    pdfproducer={Adrian Perez-Suay and Camps-Valls}, % producer of the document
    pdfkeywords={Causality}{forward}{inverse}{modelling}{dependence}, % list of keywords
    pdfnewwindow=true,      % links in new window
    colorlinks=true,       % false: boxed links; true: colored links
    linkcolor={blue},          % color of internal links (change box color with linkbordercolor)
    citecolor=blue,        % color of links to bibliography
    filecolor=blue,      % color of file links
    urlcolor=blue           % color of external links
}
\usepackage{pgffor} % for 
\renewcommand{\algorithmicrequire}{\textbf{Input:}}
\renewcommand{\algorithmicensure}{\textbf{Output:}}

\usepackage{setspace}
\newcommand{\Real}{\mathbb R}

\def\x{{\mathbf x}}

\newcommand{\vect}[1]{{\boldsymbol{\mathbf{#1}}}} % vector
\newcommand{\y}{{\vect y}}

\newcommand{\s}{{\vect s}}

\newcommand{\cut}[1]{} % cut out a part of the text

\title{Causal Inference in Geosciences with Kernel Sensitivity Maps}

\name{Adri\'an P\'erez-Suay and Gustau Camps-Valls\thanks{The research was funded by the European Research Council (ERC) under the ERC-CoG-2014 SEDAL project (grant agreement 647423), and the Spanish Ministry of Economy and Competitiveness (MINECO) through the project TIN2015-64210-R.}}{}
\address{Image Processing Laboratory (IPL), Universitat de Val\`encia, Spain}

\begin{document}

\maketitle

\begin{abstract}
Establishing causal relations between random variables from observational data is perhaps the most important challenge in today's Science. In remote sensing and geosciences this is of special relevance to better understand the Earth's system and the complex and elusive interactions between processes. In this paper we explore a framework to derive cause-effect relations from pairs of variables via regression and dependence estimation. We propose to focus on the sensitivity (curvature) of the dependence estimator to account for the asymmetry of the forward and inverse densities of approximation residuals. Results in a large collection of 28 geoscience causal inference problems demonstrate the good capabilities of the method.

\end{abstract}


\section{INTRODUCTION} \label{sec:intro}
\begin{flushright}
{\em Synoptikos, $\sigma\acute{\upsilon}\nu o \psi \iota\zeta$\\ 	
``Affording a general view of a whole.''}
\end{flushright}
Establishing causal relations between random variables from observational data is perhaps the most important challenge in today's Science. In remote sensing and geosciences this is of special relevance to better understand the Earth's system and the complex and elusive interactions between the involved processes. Answering key questions may have deep societal, economical and environmental implications~\cite{Walther02,Adam11}. 

%Most statistical methods focus on prediction or estimation problems, and are designed to study association relationships between observations and essential climate variables. While being appropriate and fruitful, this approach yields models that may lack explanatory power. It goes without saying that correlation does not imply causation. 
The purpose of {\em causal inference} is to go beyond association and to determine and discover links of {\em causes and effects}. Causal inference is performed through controlled experiments where procedures such as randomization are used to avoid selection and confounding biases. However, this is not possible in climate science and geosciences, where one cannot obviously conduct randomized experiments over the Earth's system. This is why actual experiments on the Earth-system are often replaced by numerical experiments done with an ensemble of Earth-system model simulations, a field collectively known as {\em detection and attribution}~\cite{hegerl_use_2011}. 

While a vast literature collectively perform such {\em model-based} causal inference, we will explore {\em observational-based} causal inference. % in remote sensing and geosciences, where only observational (non-experimental) data are used to infer the causal structure of the data generating system. 
Previous works in the literature most notably exploited the concept of {\em Granger's causality} to perform causal inference for the attribution of climate change~\cite{attanasio_granger_2013}, the {\em constraint-based search} to study the causal interactions between climate modes of variability~\cite{ebert-uphoff_causal_2012}, and conditional independence measures in {\em PC schemes}~\cite{pc_1993} to construct climate causal networks~\cite{runge_identifying_2015}. 

% specific setup and our proposal
In this paper we explore an alternative pathway based on regression and dependence estimation. In particular, we build upon the framework proposed in~\cite{Hoyer08} to derive cause-effect relations from pairs of variables. % via regression and dependence estimation. 
%Given two variables, the method decides on the direction of causation based on the statistical significance of the {\em dependence test} between the residuals and the input tells the most plausible causal direction, and thus indicates the true data-generating mechanism. 
The method decides about the causal direction based on the dependence of the residuals obtained after fitting a regression model in the forward and inverse directions. Authors suggested to use the Hilbert Schmidt Independence Criterion (HSIC)~\cite{Gretton05} as dependence test based on the excellent converge properties to the true dependence and ease of calculation. HSIC has been widely used in remote sensing as well for feature selection and dependence estimation~\cite{campsvalls09grsl,Camps-VallsTLM10}. 
%Note that in this approach (and in many others), causal inference strongly relies on dependence estimates. Many methods exist to this purpose. Very often one traditionally resorts to Pearson's correlation, but the measure can only identify linear associations between random variables. Other measures of dependence, such as the Spearman's rank or the Kendall's tau criteria, assume monotonically increasing variable relations, and can be better suited in problems exhibiting such relations. %All of them, however, can be computed for pairs of variables only, and thus the multidimensional problem of dependence estimation is tackled by repeating the test for all pairwise combinations, and then summarizing the `dependence matrix' into an {\em ad hoc} overall statistic. 
%Here we will focus on a kernel-based dependence estimate; the Hilbert Schmidt Independence Criterion (HSIC) introduced in~\cite{Gretton05}, which has shown excellent converge properties to the true dependence. 
We here propose to focus on the sensitivity (curvature) of the HSIC dependence estimate as it contains useful information to account for the asymmetry of the forward and inverse densities. 

The remainder of the paper is organized as follows. 
\S\ref{sec:causal} reviews the main aspects of the adopted causal framework, and the HSIC estimate.
\S\ref{sec:proposal} derives the HSIC sensitivity maps and describes the proposed causal criterion. 
\S\ref{sec:results} gives experimental evidence of performance in a large collection of 28 remote sensing and geoscience causal inference problems. 
We conclude in \S\ref{sec:conclusions} with some remarks and future outlook.

\section{Causal Inference via Regression} \label{sec:causal}

\subsection{Causality based on regression}

We build on the empirical causal approach presented in~\cite{Hoyer08} to discover causal association between variables $x$ and $y$. The methodology performs nonlinear regression from $x\to y$ (and vice versa, $y\to x$) and assesses the independence of the forward, $r_f=y-f(x)$, and backward residuals, $r_b = x-g(y)$, with the input variable $y$ and $x$, respectively. The statistical significance of the independence test tells the right direction of causation. Therefore, the framework needs two fundamental blocks: 1) a nonlinear regression model, and 2) a dependence measure. We typically rely on Gaussian Processes~\cite{CampsValls16grsm} and the HSIC~\cite{Gretton05} respectively. The final causal direction score was simply defined as the difference in test statistic between both models
$$\hat C:=\text{HSIC}(x,r_f)-\text{HSIC}(y,r_b).$$ 
%Alternatively, one may derive the HSIC $p$-values for both tests, and propose a second alternative criterion as
%$$\hat C_p:=\log(p_{x\to y})-\log(p_{y\to x}),$$
%where $p_{x\to y}$ is the $p$-value of $\text{HSIC}(x,r_f)$ and similarly for the backward direction. Either way, 
The intuition behind the approach is that statistically significant residuals in one direction indicate the true data-generating mechanism.

\subsection{Kernel Depedence Estimation with HSIC}

Let us consider two spaces ${\mathcal X}\subseteq \Real^{d_x}$ and ${\mathcal Y}\subseteq \Real^{d_y}$, on which we jointly sample observation pairs $(\x,\y)$ from distribution ${\mathbb P}_{\x\y}$. 
\iffalse
The covariance matrix can be defined as
\begin{eqnarray}\label{covmatrix}
{\mathcal C}_{\x\y} = {\mathbb E}_{\x\y}(\x\y^\top) - {\mathbb E}_{\x}(\x){\mathbb E}_{\y}(\y^\top),
\end{eqnarray}
where ${\mathbb E}_{\x\y}$ is the expectation with respect to ${\mathbb P}_{\x\y}$,  
${\mathbb E}_{\x}$ is the expectation with respect to the marginal distribution 
${\mathbb P}_{\x}$ (hereafter, we assume that all these quantities exist), and $\y^\top$ is the transpose of $\y$. 
\fi
The covariance matrix ${\mathcal C}_{\x\y}$ encodes first order dependencies between the random variables. A statistic that efficiently summarizes the content of this matrix is its Hilbert-Schmidt norm. 
\iffalse
The square of this norm is equivalent to the squared sum of its eigenvalues $\gamma_i$:
\begin{eqnarray}
\|{\mathcal C}_{\x\y}\|_{\text{HS}}^2 = \sum_i \gamma_i^2.
\end{eqnarray}
\fi
This quantity is zero if and only if there exists no first order dependence between $\x$  and $\y$. Note that the Hilbert Schmidt norm is limited to the detection of second order relations, and thus more complex (higher-order effects) cannot be captured.

%%% kernels....

The nonlinear extension of the notion of covariance was proposed in \cite{Gretton05}. Let us define a (possibly non-linear) mapping $\boldsymbol{\phi}: {\mathcal X}\to {\mathcal F}$ such that the inner product between features is given by a positive definite (p.d.) kernel function  $K_x(\x,\x') = \langle \boldsymbol{\phi}(\x),\boldsymbol {\phi}(\x')\rangle$. The feature space ${\mathcal F}$ has the structure of a reproducing kernel Hilbert space (RKHS). Let us now denote another feature map $\boldsymbol{\psi}:{\mathcal Y}\to {\mathcal G}$ with associated p.d. kernel function $K_y(\y,\y') = \langle \boldsymbol{\psi}(\y),\boldsymbol{\psi}(\y')\rangle$. Then, the cross-covariance operator between these feature maps is a linear operator ${\mathcal C}_{\x\y}:{\mathcal G} \to {\mathcal F}$ such that
$
{\mathcal C}_{\x\y} = 
%{\mathbb E}_{\x\y}[(\boldsymbol{\phi}(\x)-\mu_x)(\boldsymbol{\psi}(\y)-\mu_y)^\top],
{\mathbb E}_{\x\y}[(\phi(\x)-\mu_x)\otimes(\psi(\y)-\mu_y)],
$
where $\otimes$ is the tensor product, $\mu_x={\mathbb E}_{\x}[\boldsymbol{\phi}(\x)]$, and $\mu_y={\mathbb E}_{\y}[\boldsymbol{\psi}(\y)]$. 
%, and ${\bf u}{\bf v}^\top$ denotes the linear operator ${\bf u}{\bf v}^\top: {\mathcal G} \to {\mathcal F}$, ${\bf w}\mapsto {\bf u} \langle {\bf v},{\bf w}\rangle$. 
See more details in \cite{Baker73,Fukumizu04}. The squared norm of the cross-covariance operator, $\|{\mathcal C}_{\x\y}\|^2_{\text{HS}}$, is called the Hilbert-Schmidt Independence Criterion (HSIC) and can be expressed in terms of kernels~\cite{Gretton05}.
%\begin{eqnarray}
%\begin{array}{lll}
%\text{HSIC}({\mathcal F},{\mathcal G},{\mathbb P}_{\x\y}) & 
%= & \|{\mathcal C}_{\x\y}\|^2_{\text{HS}} \nonumber\\[2mm]
%& = &{\mathbb E}_{\x\x'\y\y'}[k_x(\x,\x')k_y(\y,\y')] \\[2mm]
%& & + {\mathbb E}_{\x\x'}[k_x(\x,\x')]{\mathbb E}_{\y\y'}[k_y(\y,\y')]  \nonumber\\ [2mm]
%& & - 2{\mathbb E}_{\x\y}[{\mathbb E}_{\x'}[k_x(\x,\x')] {\mathbb E}_{\y'}[k_y(\y,\y')]], \nonumber
%\end{array}
%\end{eqnarray}
%where ${\mathbb E}_{\x\x'\y\y'}$ is the expectation over both $(\x,\y) \sim {\mathbb P}_{\x\y}$ and an additional pair of variables $(\x',\y') \sim {\mathbb P}_{\x\y}$ drawn independently according to the same law.
%An empirical estimate of HSIC can be easily obtained. 
Given the sample datasets ${\bf X}\in\Real^{n\times d_x}$, ${\bf Y}\in\Real^{n\times d_y}$, 
%and the joint dataset ${\mathcal Z}={\mathcal X}\times {\mathcal Y}$, 
with $n$ pairs $\{(\x_1,\y_1),\ldots,(\x_n,\y_n)\}$ 
%collectivelly grouped in a sample matrix ${\bf Z}\in\Real^{n\times(d_x+d_y)}$ 
drawn from the joint ${\mathbb P}_{\x\y}$, 
an empirical estimator of HSIC is~\cite{GreBouSmoSch05}:
\begin{eqnarray}
\text{HSIC}({\mathcal F},{\mathcal G},{\mathbb P}_{\x\y}) = %\dfrac{1}{n^2} \text{Tr}({\bf K}_x{\bf H}{\bf K}_y{\bf H}) = 
\dfrac{1}{n^2} \text{Tr}({\bf H}{\bf K}_x{\bf H}~{\bf K}_y),
\label{empHSIC}
\end{eqnarray}
where $\text{Tr}(\cdot)$ is the trace operation (the sum of the diagonal entries),  ${\bf K}_x$, ${\bf K}_y$ are the kernel matrices for the input random variables $\x$ and $\y$, respectively, and ${\bf H}= {\bf I} - \frac{1}{n}\mathbbm{1}\mathbbm{1}^\top$ centers the data in the feature spaces ${\mathcal F}$ and ${\mathcal G}$, respectively. %Here $\delta$ represents the Kronecker symbol, where $\delta_{i,j}=1$ if $i=j$, and zero otherwise. 

\section{Causal Inference with Sensitivity Maps} \label{sec:proposal}

HSIC has demonstrated excellent capabilities to detect dependence between random variables but, as for any kernel method, the learned relations are hidden behind the kernel feature mapping. %Visualization and geometrical interpretation of kernel methods in general and kernel dependence estimates in particular is an important issue in machine learning. 
To address this issue, we next derive sensitivity maps for HSIC that account for the relevance of the features and the examples in the measure. Then, we use these sensitivity maps to propose a novel criterion for causal inference.

\subsection{Sensitivity analysis and maps}

A general definition of the {\em sensitivity map} was originally introduced in~\cite{Kjems02}. The sensitivity map is somewhat related to the concepts of {\em leveraging} and {\em influential observations} in statistics~\cite{burt2009elementary}. Let us define a function $f(\param):\Real^d\to\Real$ parametrized by $\param=[z_1,\ldots,z_d]$. The sensitivity map for the $j$th feature, $z_j$, is the expected value of the squared derivative of the function (or the log of the function) with respect the arguments. Formally, let us define the sensitivity of $j$th feature as 
\begin{equation}
s_j=\int_{\mathcal Z}\bigg(\dfrac{\partial f(\param)}{\partial z_j}\bigg)^2p(\param)\text{d}\param,
\label{smeander}
\end{equation}
where $p(\param)$ is the probability density function over the input $\param\in{\mathcal Z}$. Intuitively, the objective of the sensitivity map is to measure the changes of the function $f(\param)$ in the $j$th direction of the input space. 
In order to avoid the possibility of cancellation of the terms due to its signs, the derivatives are typically squared, even though other unambiguous transformations like the absolute value could be equally applied. Integration gives an overall measure of sensitivity over the observation space ${\mathcal Z}$. The {\em empirical sensitivity map} approximation to Eq.~\eqref{smeander} is obtained by replacing the integral with a summation over the available $n$ samples 
\begin{equation}\label{empirical}
s_j\approx\dfrac{1}{n}\sum_{i=1}^n \dfrac{\partial f(\param)}{\partial z_j}\bigg|_{\param_i}^{~2},
\end{equation}
which can be collectively grouped to define the {\em sensitivity vector} as $\s:=[s_1,\ldots,s_d]$. 

\subsection{Sensitivity maps for the HSIC}

The previous definition of SMs is limited in some aspects: 1) it cannot be directly applied to general functions $f$ depending on matrices or tensors, and 2) it allows estimating feature relevances $s_j$ only, discarding the individual samples relevance. The former is a severe limitation in particular for HSIC as it directly deals with data matrices. % The latter limitation affects all type of functions including HSIC, which could be useful in applications on outlier detection or sample selection, just to name a few. 
In the sequel we derive the HSIC with respect to both the input samples and features simultaneously to address both shortcomings. 

Let us now define $f:=\text{HSIC}$, and replace the gradient $\frac{\partial f(\param)}{\partial z_j}$ with $\frac{\partial f(\Param)}{\partial Z_{ij}}$. Note that point-wise HSIC is parametrized by $\param\sim \x,\y|\sigma$, while a matrix parametrization reduces to $\Param\sim{\bf X},{\bf Y}|\sigma$. In order to compute the HSIC sensitivity maps, we derive HSIC w.r.t. input data matrix entries $X_{ij}$ and $Y_{ij}$. By applying the chain rule, and first-order derivatives of matrices, we obtain:
%Now, deriving the HSIC estimate in Eq.~\eqref{empHSIC} w.r.t. every $i$-th example and $j$-th feature of $\X$ and $\Y$, it is easy to find: %$${\bf s}_j^x:=\dfrac{\partial\text{HSIC}}{\partial{\bf x}^r}=-\frac{1}{\sigma^2(n-1)^2}{\bf K}_x{\bf H}{\bf K}_y{\bf H}\sum_{k=1}^d\delta_r^k(x_i^k-x_j^k),$$ 
\begin{equation}\label{shsic_x}
%S_{ij}^x:=\dfrac{\partial\text{HSIC}}{\partial X_{ij}}=-\frac{2}{\sigma^2 n^2}\text{Tr}\left({\bf H}{\bf K}_y{\bf H}({\bf K}_x\circ {\bf M}_{ik}^j)\right),
S_{ij}^x:=\dfrac{\partial\text{HSIC}}{\partial X_{ij}}=-\frac{2}{\sigma^2 n^2}\text{Tr}\left({\bf H}{\bf K}_y{\bf H}({\bf K}_x\circ {\bf M}_j\right),
\end{equation} 
where for a given $j$th feature, entries of the corresponding matrix ${\bf M}_j$ are $M_{ik}=X_{ij}-X_{kj}$ ($1\leq i,k\leq n$), and the symbol $\circ$ is the Hadamard product between matrices. After operating similarly for $Y_{ij}$, we obtain the corresponding expression:
\begin{equation}\label{shsic_y}
S_{ij}^y:=\dfrac{\partial\text{HSIC}}{\partial Y_{ij}} = -\frac{2}{\sigma^2 n^2}\text{Tr}\left({\bf H}{\bf K}_x{\bf H}({\bf K}_y\circ {\bf N}_j)\right),
\end{equation}
where entries of matrix ${\bf N}_j$ are $N_{ik}=Y_{ij}-Y_{kj}$, and we assumed the SE kernel for both variables\footnote{Other positive definite universal kernels could be equally adopted, but that requires deriving the sensitivity maps for the specific kernel.} % In addition, as seen in Section~\ref{sec:rhsic}, random features approximation works for shift invariant kernels only.}. 

It is worth noting that, even though one could be tempted to use each sensitivity map independently, the solution is a {\em vector field}, and we should treat the sensitivity map jointly. To do this we define the {\em total sensitivity map} for all features and samples as ${\bf S}:=[{\bf S}^x,{\bf S}^y]\in\Real^{n\times (d_x+d_y)}$. From ${\bf S}$ we can compute the {\em empirical sensitivity map} per either feature or sample by integration in the corresponding domain, whose empirical estimates are respectively 
$s_i\approx\frac{1}{d}\sum_{j=1}^d S_{ij}^2$ and $s_j\approx\frac{1}{n}\sum_{i=1}^n S_{ij}^2.$
%\begin{equation}\label{empirical2}
%s_i\approx\dfrac{1}{d}\sum_{j=1}^d \dfrac{\partial \text{HSIC}}{\partial Z_{ij}}\bigg|_{\Param_j}^{~2},~~\text{and}~~
%s_j\approx\dfrac{1}{n}\sum_{i=1}^n \dfrac{\partial \text{HSIC}}{\partial Z_{ij}}\bigg|_{\Param_i}^{~2}.
%\end{equation}
This complementary view of the sensitivity reports information about the directions most impacting the dependence estimate, and allows a quantitive geometric evaluation of the measure. 

\subsection{Proposed criterion}

We here propose an alternative criterion for association based on the sensitivity of HSIC in both directions:
$$\hat C_s:=(S_b^y + S_b^{r})-(S_f^x + S_f^{r}),$$ 
where subscripts $f$ and $b$ stand for the forward and backward directions, and the superscripts refer to the sensitivities of either the observations $x$ and $y$, or the corresponding residuals, $r_f$ and $r_b$. Similar criteria has been recently presented in the literature, yet focusing on the derivative of the underlying function, instead of the derivative of the dependence estimate itself~\cite{Daniusis12}.

\section{RESULTS} \label{sec:results}

%We show experimental results in two annotated datasets of cause-effect pairs related to geoscience and remote sensing problems. This will allow us to quantify the accuracy in detecting the direction of causation using standard scores.
We show experimental results in an annotated dataset of 28 cause-effect pairs related to geoscience and remote sensing problems. This will allow us to quantify the accuracy in detecting the direction of causation using standard scores.

%\subsection{Experiment 1: Geoscience Cause Effect Pairs}
\subsection{Geoscience Cause Effect Pairs}

\begin{figure}[t!]  % 1:100,[52:55 71]  foreach
\foreach \i in {1,2,3,4,20,21,42,43,44,45,46,49,50,51,72,73,78,79,80,81,82,83,87,89,90,91,92,93} {
\includegraphics[width=1cm]{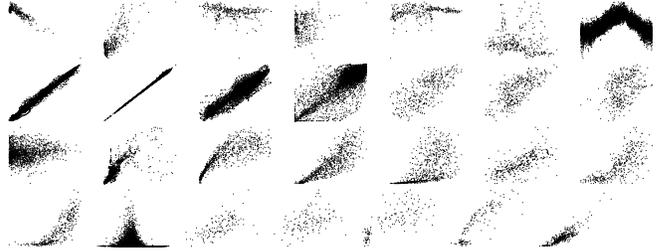}
}
\caption{Scatter plots in the CEP benchmark. \label{fig:subfig}}
\end{figure}

We used Version 1.0 of the CauseEffectPairs (CEP) collection\footnote{\url{https://webdav.tuebingen.mpg.de/cause-effect/}}. The database contains 100 pairs of random variables along with the right direction of causation (ground truth). Data has been collected from various domains of application, such as biology, climate science and economics, just to name a few. More information about the dataset and an excellent up-to-date review of causal inference methods is available in~\cite{Mooij16}. We conducted experiments in 28 out of the 100 pairs that have one-dimensional variables and that are related to geosciences and remote sensing: problems involving carbon and energy fluxes, ecological indicators, vegetation indices, temperature, moisture, heat, etc. Scatter plots of the selected pairs $x-y$ are shown in Fig.~\ref{fig:subfig}. 

\subsection{Experimental Setup}
The experimental setting is as follows. We used random forests (RFs)\footnote{We also tried Gaussian process regression as in~\cite{Mooij16} but results were less accurate and more computationally demanding.} for both forward and backward regression models, and then computed HSIC and the sensitivity of HSIC. The final causal direction score was simply defined as the difference in test statistic between both models, either using $\hat C$ and the proposed $\hat C_s$. Evaluation of the results needs to quantify the accuracy in detecting the causal direction. Note that this is a particular form of `ranked-decision' setting that needs to account for the bias introduced by pairs coming from the same problem, i.e. it is customary to down-weight the precision for every decision threshold in the curves (e.g. four related problems receive 0.25 weights in the decision)\footnote{The MATLAB function \url{perfcurve} can produce such (weighted) ROC and PRC curves and the estimated weighted AUC.}.

\subsection{Accuracy and robustness of the detection}
\begin{figure}[h!]
\setlength{\tabcolsep}{1pt}
\begin{center}
\begin{tabular}{cccc}
\includegraphics[width=4.2cm]{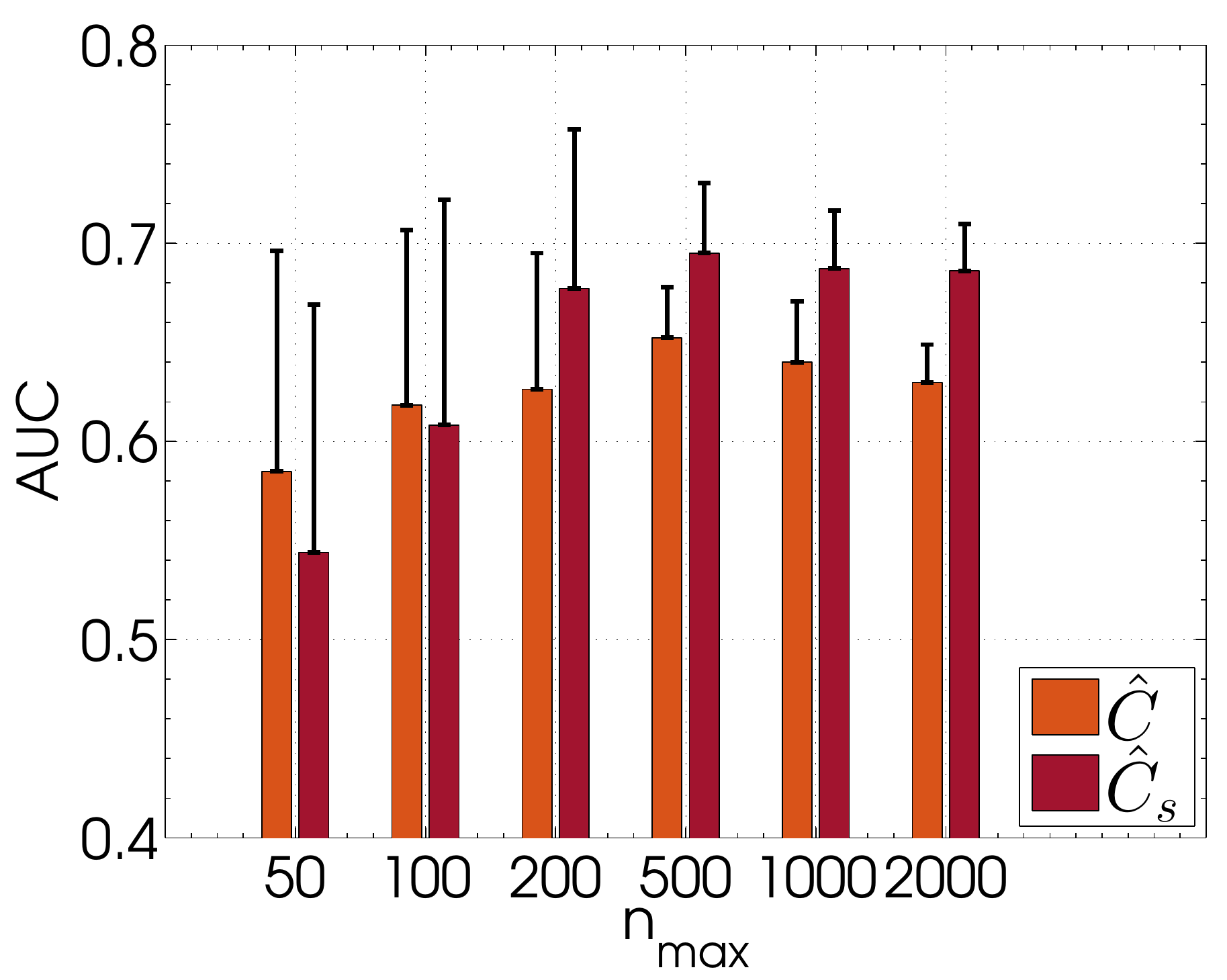}  & 
\includegraphics[width=4.2cm]{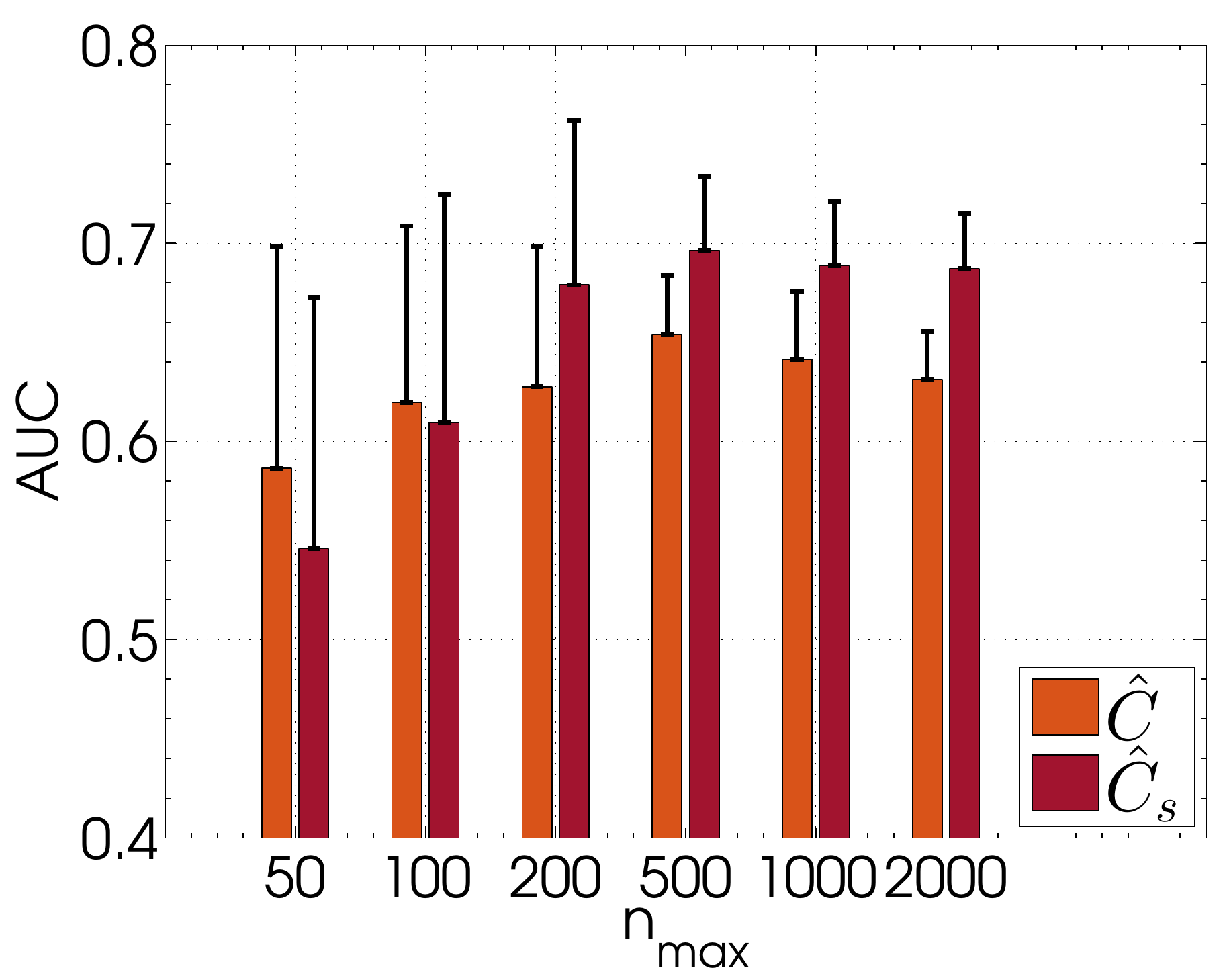}  & 
\end{tabular}
\end{center}
\vspace{-0.7cm}
\caption{ROC and corresponding AUCs in the CEP causality problems dataset for $n_{max}=200$ (thin lines) and $n_{max}=2000$ (thick lines) in every problem. \label{fig:sensrocbars}}
\end{figure}

We run the experiments with different numbers of (randomly selected) points $n$ from both random variables. This situation dramatically impacts regression models performance, both in terms of the regression accuracy and the dependence estimation. We evaluate $\hat C$ and $\hat C_s$ criteria by limiting the maximum number of training samples in each problem, $n_{max}=\{50,100,200,500,2000\}$. Results were averaged over $10$ realizations. Figure~\ref{fig:sensrocbars} shows the ROCs for both criteria, and the AUC under the curves as a function of $n_{max}$. The proposed sensitivity-based criterion consistently performed better than the standard approach using HSIC alone. 

\begin{figure}[h!] 
\setlength{\tabcolsep}{1pt}
\begin{center}
\begin{tabular}{cc}
\includegraphics[width=4.2cm]{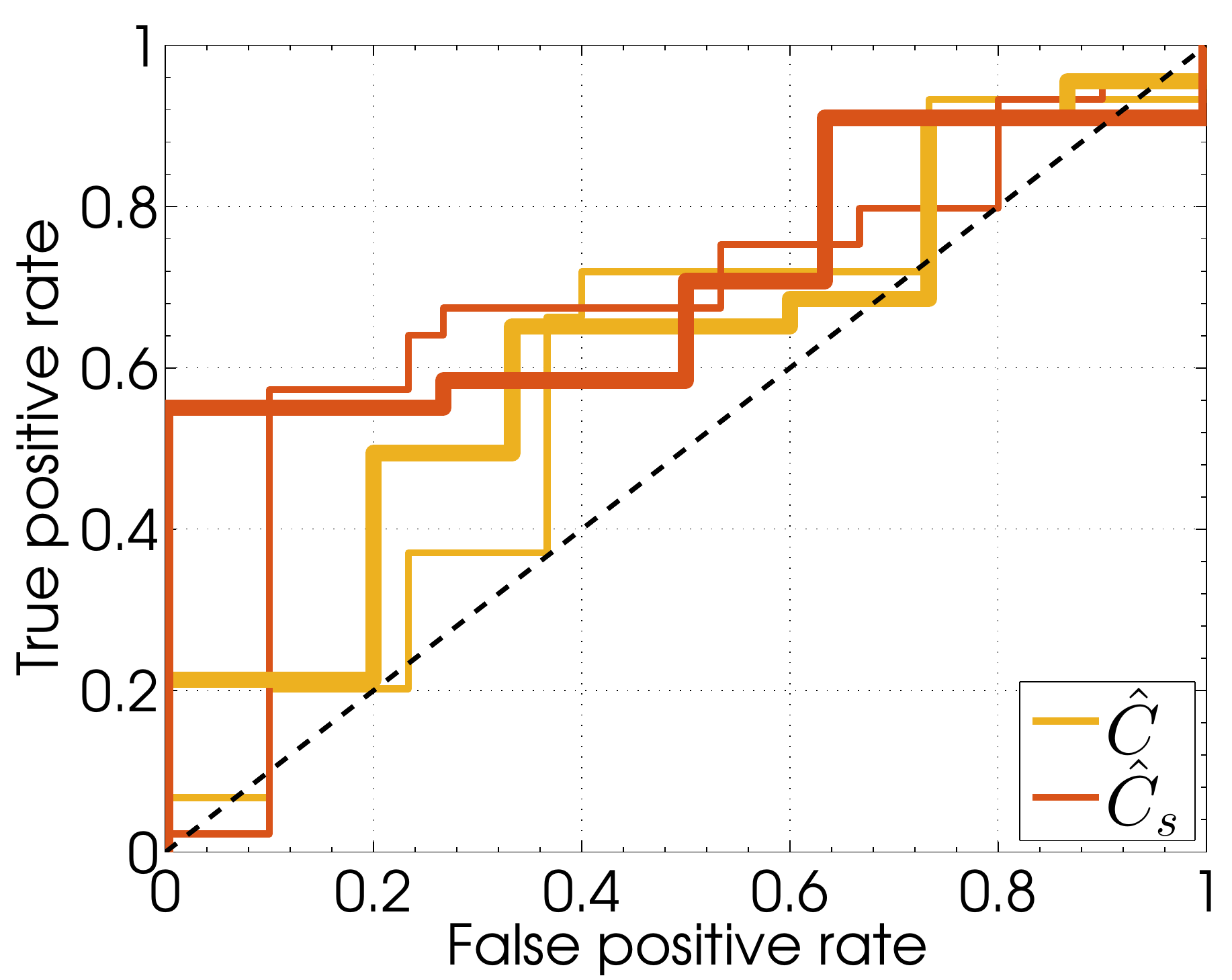} &
\includegraphics[width=4.2cm]{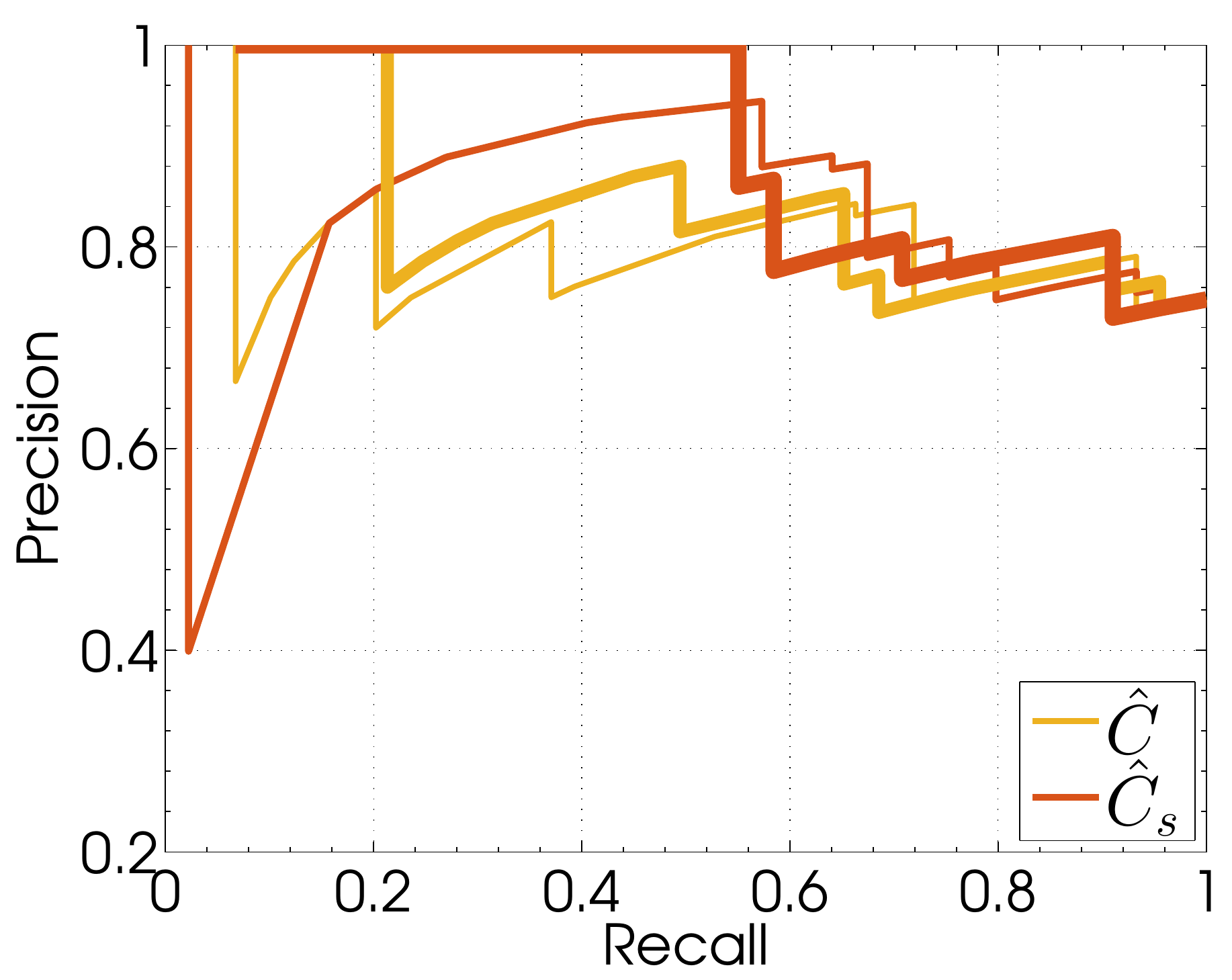} 
\end{tabular}
\end{center}
\caption{ROC (left) and Precision-Recall (right) curves for the causality problem using $n=\{50,200,2000\}$ samples (thicker lines mean higher $n$. \label{fig:roc}}
\end{figure}

The best recognition curves for both criteria are given in Fig. \ref{fig:roc}. It can be noted that in both cases, the ROC and the precision recall curves (PRC) for the proposed $\hat C_s$ are better than those of $\hat C$, and this happens for all decision rates, especially in the false positive rates regime.

\section{CONCLUSIONS} \label{sec:conclusions}
We introduced a simple method based on observational data to uncover cause-effect relations in geoscience and remote sensing problems. We relied on a framework based on regression and dependence estimation, and proposed to focus on the sensitivity (curvature) of the dependence estimator instead of the dependence itself. This allows us to better capture the asymmetry of the forward and inverse densities. Results in a large collection of 28 geoscience causal inference problems demonstrated the good performance of our proposal.

% REFERENCES
\small
%\nocite{*}
\bibliographystyle{IEEEbib}
\bibliography{DSPKM,random,cites,ei}

\end{document}